\documentclass[letterpaper]{article} 
\usepackage[submission]{aaai24}  
\usepackage{times}  
\usepackage{helvet}  
\usepackage{courier}  
\usepackage[hyphens]{url}  
\usepackage{graphicx} 
\usepackage{natbib}  
\usepackage{caption} 
\usepackage{algorithm}
\usepackage{algorithmic}
\usepackage{xspace}

\usepackage{newfloat}
\usepackage{listings}
\DeclareCaptionStyle{ruled}{labelfont=normalfont,labelsep=colon,strut=off} 
\floatstyle{ruled}
\newfloat{listing}{tb}{lst}{}
\floatname{listing}{Listing}
\usepackage{amsmath}
\usepackage{dsfont}
\usepackage[inline]{enumitem}
\usepackage{amsfonts}
\newtheorem{definition}{Definition}
\usepackage{multirow}
\usepackage{booktabs}
\usepackage{makecell}
\usepackage[capitalize,nameinlink]{cleveref}
\newcommand*{\defeq}{\stackrel{\text{def}}{=}}
\newcommand*{\eqrepa}{\stackrel{\text{REPA}}{=}}

\title{\acronym: Client Clustering without Training and Data Labels for Improved Federated Learning in Non-IID Settings}

\author {
    Boris Radovič\textsuperscript{\rm 1},
    Veljko Pejović\textsuperscript{\rm 1,\rm 2}
}
\affiliations {
    \textsuperscript{\rm 1}Faculty of Computer and Information Science, University of Ljubljana, Slovenia\\
    \textsuperscript{\rm 2}Lancaster University, UK
}

\usepackage{bibentry}

\newcommand{\acronym}{REPA\xspace}

\begin{document}

\maketitle

\begin{abstract}

Clustering clients into groups that exhibit relatively homogeneous data distributions represents one of the major means of improving the performance of federated learning (FL) in non-independent and identically distributed (non-IID) data settings. Yet, the applicability of current state-of-the-art approaches remains limited as these approaches cluster clients based on information, such as the evolution of local model parameters, that is only obtainable through actual on-client training. On the other hand, there is a need to make FL models available to clients who are not able to perform the training themselves, as they do not have the processing capabilities required for training, or simply want to use the model without participating in the training. Furthermore, the existing alternative approaches that avert the training still require that individual clients have a sufficient amount of labeled data upon which the clustering is based, essentially assuming that each client is a data annotator. In this paper, we present \acronym, an approach to client clustering in non-IID FL settings that requires neither training nor labeled data collection. \acronym uses a novel supervised autoencoder-based method to create embeddings that profile a client's underlying data-generating processes without exposing the data to the server and without requiring local training. Our experimental analysis over three different datasets demonstrates that \acronym delivers state-of-the-art model performance while expanding the applicability of cluster-based FL to previously uncovered use cases.





\end{abstract}

\section{Introduction}
\label{sec:introduction}

Federated learning (FL) \cite{fl_seminal_paper} is a distributed machine learning paradigm in which a model is trained locally over a set of clients, with only the model updates being observed and aggregated by a server. FL ensures that the clients' data remain on the devices and offloads the training burden from the server, thus representing an important step towards decentralized AI.

Avoiding aggregating the data on the server to ensure privacy, on the other hand, hinders the model convergence in cases of diverging data distributions across clients that participate in FL. The problem of non-independent and identically distributed (non-IID) data is known to hamper the training convergence and the accuracy of the final model~\cite{loooong}. Consequently, approaches have been proposed to go beyond FedAvg \cite{fl_seminal_paper} -- a commonly used simple aggregation of local model updates. Among these approaches, of particular interest are those that rely on grouping clients into clusters exhibiting relatively homogeneous data distributions, such as~\cite{clustering_fesem,clustering_hierarchical}. After clustering the clients the server can evolve different (flavors of) models for different clusters of clients. Such ensembles of models can improve convergence and inference accuracy in non-IID settings, while still ensuring that the clients' data remains fully private.

Unfortunately, real-world applicability of clustering-based solutions to FL in non-IID settings remains limited, since most clustering methods rely on information stemming from local training. This information, such as the direction in which the model parameters are evolving during local training, is only obtainable by clients who have sufficient computational and energy resources to conduct on-device training. In real-world applications, however, FL-trained models will often be used by clients who may never get a chance to participate in the training themselves. For instance, Android's predictive on-screen keyboard -- Gboard -- is trained in a federated manner, yet in each round, only a few hundred high-end smartphones connected to a charger participate in the training~\cite{bonawitz2019towards}. Consequently, comprehensive clustering that would allow personalized model deployment across billions of Android users remains infeasible, if training is mandated for each client. While alternative approaches that alleviate the need for training have appeared~\cite{clustering_send_all_2_all,ruan2022fedsoft}, they still require that each client collects and labels a substantial amount of data upon which the clustering is performed. Such a requirement implies that each client has to serve as a data annotator, even if only interested in using, not necessarily training, the model.

In this paper, we present \acronym, a novel method for client clustering in non-IID data settings. Unlike the existing solutions, \acronym brings benefits of personalized FL modeling to a wide range of clients, including those who have no labeled data and those who cannot train, either temporarily, e.g. due to insufficient battery power, or at all, e.g. due to poor hardware capabilities. To enable this, \acronym develops an encoder to embed the client's unlabeled data into a latent space. The clustering is then performed on profiles of the embedded data. We show that statistics calculated over the generated data embeddings reflect the actual data distribution sufficiently well to allow robust clustering. Moreover, experimenting with different datasets we demonstrate that \acronym matches the state-of-the-art inference in FL while making customized models available to a wide range of training-free clients. Specifically, the contributions of this paper include:
\begin{itemize}
    \item We devise a supervised autoencoder-based embedder through which we profile the data without the need for training and without the need for labeled data;
    \item We propose a lightweight FL client clustering scheme that harnesses statistical profiles of the embedding distributions. Neither the data nor the actual data distributions need to be sent to the server;
    \item We compare the performance of \acronym against the state-of-the-art scheme for client clustering that relies on on-client training, and demonstrate that our approach delivers comparable performance while significantly expanding the usability of FL-trained models.
\end{itemize}

\section{Related Work}
\label{sec:related_work}

The stochastic gradient used for FL training is an unbiased estimate of the full gradient only if gradients produced by individual clients are all calculated over data samples originating from the equivalent data-generating processes (DGPs)~\cite{zhao2018federatednoniid}. In practice, however, DGPs may vary across clients, for instance, due to the DGPs' tight relationship to individual human behavior, including one's movement patterns, keyboard typing habits, voice properties, and other factors. In case of significant non-IIDness, straightforward aggregation of local gradients at the server may lead to drastically underperforming FL models, with studies demonstrating a 55\% reduction in model accuracy on the keyword spotting and 51\% accuracy loss on CIFAR-10 image recognition tasks~\cite{zhao2018federatednoniid}.

Individual clients ``pulling'' the global model in different directions prevent model convergence in FL. To counter this, Li et al. propose penalizing the deviation of local models from the global model through L2 regularization of model weights~\cite{fedprox}. Alternatively, contrastive learning can be harnessed to pull the embeddings of data instances belonging to the same classes close to each other in the embedding space, and the embeddings of data instances belonging to different classes away from each other~\cite{fedproc}. This is done across all clients, who are orchestrated to align these embeddings with class prototypes calculated at the global level.

However, even if the above solutions manage to improve the convergence, a single global model may not perform well across the range of clients that train and/or use it. In such cases, clustering clients into groups exhibiting similar DGPs might be more appropriate, as it would allow for multiple model flavors to be trained -- one for each cluster of clients. Nevertheless, such clustering is challenging in FL settings, as it has to satisfy what appear to be conflicting requirements: reflect individual clients' data distributions and maintain the clients' data privacy. Consequently, clustering solutions have appeared that rely on data that is an inherent part of federated training. The approach presented in \cite{clustering_hierarchical}, for example, first trains the joint model $\theta^T$ on all the available clients up to round $T$ and then computes the difference between the joint model and each locally evolved model $\Delta \theta_i=\theta^T - \theta_i^{T+1}$ for every client. This vector is then used for hierarchically clustering the clients. Similarly, FeSEM~\cite{clustering_fesem} randomly initializes multiple models and iteratively assigns clients to a cluster whose model parameters are nearest to the client's locally-tuned model according to a certain distance measure. Other algorithms leverage information directly related to the gradient updates to cluster the clients as well \cite{duan2021fedgroup,clustered_fl_first}.


The above approaches rely on information accessible only to clients capable of training a model. Consequently, a potentially large number of clients unable to perform this action cannot be assigned to a cluster and, as a result, do not obtain the corresponding model flavor. Iterative Federated Clustering Algorithm \cite{clustering_send_all_2_all} avoids this training requirement by letting the clients determine which cluster they belong to themselves. To do so, each client calculates the value of a certain loss function over its local dataset for all the model flavors and selects the model flavor that results in the lowest loss. For calculating the loss, however, this approach requires that each client collects a labeled dataset. Obtaining data labels is known to be expensive in many mobile sensing applications and avoiding this restriction would greatly improve the solution generalizability.

To the best of our knowledge, the only clustering solution that requires neither labeled data nor on-device training is \cite{health_autoencoder}. The approach, however, clusters individual data points, not individual clients. Embeddings of data points obtained through an autoencoder are clustered, and a separate model is trained for every cluster. Running multiple instances of FedAvg training in parallel across clients is feasible as the algorithm targets cross-silo FL settings, where the training is performed by a small number of relatively reliable clients with large datasets, for instance, hospitals. Nevertheless, the approach is unlikely to be suitable for resource-constrained devices, such as smartphones, and dynamic ubiquitous computing settings.

\section{Preliminaries and General Framework for Clustered FL}
\label{sec:methods}

The related work analysis from the previous section reveals no unified framework for considering clustering FL clients in non-IID settings. Such a framework is necessary should we wish to ensure that competing clustering solutions are evaluated on the same terms. Thus, in this section, we introduce a general framework for clustered FL.

We assume that the set of clients $C$ can be partitioned as follows:
\begin{itemize}
    \item $C_t\subseteq C$ \textit{training clients}, which are able to train a model on their local datasets;
    \item $C_h\subset C$ \textit{holdout clients}, which wish to use an FL-trained model, yet, do not participate in model training, either because of the lack of local labeled datasets, insufficient computing capabilities, or other reasons;
\end{itemize}
and that $C_t \cup C_h = C$ and $C_t \cap C_h=\emptyset$.

Further, we assume that each client $i\in C$ possesses a (not necessarily labeled) dataset $D_i$ generated by sampling data from the single Data-Generating Process (DGP) $\gamma_i$ associated with the client. That is,
\begin{align}
\label{eq:dgp}
D_i=\{X_{i,j}, y_{i, j}\}_{j=1}^{|D_i|}\sim_{i.i.d.} \gamma_i.
\end{align}

The concept of DGP is abstract, as it represents an unknown stochastic process that encompasses all factors influencing the collected data. From a probabilistic perspective, any DGP $\gamma_i$ is characterized by a distribution $p_i(x, y)$, which may be rewritten as $p_i(x)\cdot p_i(y|x)$ or $p_i(y)\cdot p_i(x|y)$\footnote{For simplicity we use the shorthand notation $p(x)=p_X(x)$, $p_{X|Y}(x|y)=p(x|y)$, etc.}.  If any of these probabilities differ between two DGPs, we say that the data generated by the two DGPs is non-IID. In particular, the non-IIDness types may be characterized as follows \cite{loooong}:

\begin{itemize}
    \item \textit{Concept drift}, when $p_i(x|y)\neq p_j(x|y)$;
    \item \textit{Concept shift}, when $p_i(y|x)\neq p_j(y|x)$;
    \item \textit{Label skew}, when $p_i(y)\neq p_j(y)$;
    \item \textit{Feature skew}, when $p_i(x)\neq p_j(x)$;
\end{itemize}

The complexity of DGPs prevents their straightforward use in client clustering. Thus, we introduce the notion of \textit{client embedding} $e_i=F(\gamma_i)$, a semantically rich vector that summarizes the properties of a client $i$'s DGP $\gamma_i$. Since these properties often remain unknown, clients need to compute their embeddings by leveraging solely their local dataset $D_i$, the data sent from the server $S$, and possibly other contextual information $\kappa_i$ they have at their disposal:

\begin{align}
    e_i=F(\gamma_i) \approx f(D_i, S, \kappa_i)
\end{align}

The goal of clustering is to assign clients with similar DGPs to the same cluster so that a personalized model can be grown within each cluster. The clustering can be performed directly on embeddings, as these embeddings reflect the DGPs. Since holdout clients may request the model after the training has been completed, the algorithm used for clustering must be able to handle embeddings that were not seen during the training. Thus, the K-Means clustering algorithm \cite{kmeans} is a viable choice, while DBSCAN \cite{dbscan} is not.

To summarize, by clustering the clients we transform the training task from training a single model $\theta$ in conditions of data heterogeneity to the training of multiple models $\theta_k$, each in a more homogeneous environment. Consequently, instead of minimizing the loss of a single global model across all clients, the training task boils down to training one model $\theta_k$ for every cluster of clients:

\begin{align}
    \theta_k^*=\arg \min_{\theta\in \mathbb{R}^d} \,\, \frac{1}{\sum_{i\in C} \mathds{1}_{\alpha(i)=k}}\sum_{i=1}^{|C|} \mathds{1}_{\alpha(i)=k} \cdot L_i(\theta),
\end{align}
where $L_i(\theta)=\mathbb{E}_{(x,y)\sim \gamma_i} [l(\theta, x; y)]$ is the expected loss over client $i$'s DGP of the model parametrized with $\theta$, $\mathds{1}_{\alpha(i)=k}$ is the indicator function taking value $1$ if $\alpha(i)=k$ else $0$, and $\alpha$ a function that assigns every client to a cluster. The function $\alpha$ logically consists of two steps: first, it computes the client embeddings of every available client and second, it uses such client embeddings to fit a clustering model and hence partition the clients.

After clients are clustered, training can proceed in a traditional FL way, i.e. at each server epoch $t$ every client $i$ selected for training receives the cluster model $\theta_{\alpha(i)}^t$, finetunes it so as to obtain the model $\theta_{\alpha(i), i}^{t+1}$, and returns it to the server. The latter then computes the new version of the cluster model as:

\begin{align}
    \theta_{k}^{t+1} = W\Big(\bigcup_i\big\{ (\theta_{\alpha(i), i}^{t+1}, \epsilon_i)~ | ~ \alpha(i)=k \big\} \Big),
\end{align}
where $W$ is the weight aggregation function and $\epsilon_i$ is the meta-information sent by the clients, e.g. the client's training dataset size. When using the FedAvg algorithm, the function $W$ simply computes the weighted average of the received model parameters.

\subsection{\acronym}
\label{sec:client_embedding_models}

We now present \acronym, a novel algorithm for computing client embeddings that \textit{does not involve any training} on the client's datasets and that \textit{can also be used for computing the embeddings of clients that do not possess a labeled dataset}, i.e. whose $y_{i, j}$ in \Cref{eq:dgp} are unknown. We first briefly discuss the rationale behind \acronym.

Aiming to devise an algorithm that may be used for computing embeddings of clients who do not possess a labeled dataset, we necessarily need to restrict ourselves to a set $\{X_{i, j}\}_{j=1}^{|D_i|}$. Such a set is, from a probabilistic perspective, a collection of random variables following the distribution $p_i(x)$. Note, that $p_i(x)$, when mapped with a function $\phi$ to another space, induces a probability distribution over that space\footnote{This holds only if the function $\phi$ is Borel measurable.}. Although $p_i(x)$ is unknown, similar to the other distributions in the DGP, the available data forms an empirical distribution which, as the number of samples $|D_i|$ increases, increasingly well approximates the true underlying distribution -- this is formally defined by the limit theorems.

Input data points might be of very high dimensionality and not semantically rich enough,  thus, our function $\phi$ harnesses deep learning to map input data to a lower dimensional and semantically richer embedding space $\mathbb{R}^{E}$. The empirical distribution of the original DGP, when mapped to the embedding space with a Borel measurable function, induces an empirical distribution over the embedding space. We can summarize the properties of the embedding space by computing a number of statistics and concatenating them into a single vector with a function $\mathbf{S}$. Such statistics might include the mean vector, the cross-covariance matrix, the cross-correlation matrix, the principal components, or any statistics over the marginal distributions, e.g. the skewness, the kurtosis, and quantiles.

To summarize, our hypothesis, formalized in \Cref{eq:esc_reasoning}, is that statistics computed over the embedding space reasonably well reflect the DGPs. As we cannot say a priori which statistics best summarize the properties of the unknown embedding space, we identify such statistics experimentally later in the paper.
\begin{align}
\label{eq:esc_reasoning}
    e_i=F(\gamma_i)\approx f(\mathcal{D}_i, \kappa_i, S) \eqrepa \mathbf{S}(\phi(\{X_{i,j}\}_{j=1}^{|\mathcal{D}_i|}))
\end{align}

The function $\phi$ used to map the individual data points $X_{i, j}$ to the embedding space might be a pre-trained encoder network or the encoder part of a network trained with FL. There are various ways to train such an encoder. For instance, through a self-supervised task with clients jointly training an autoencoder in an FL setting. Or, for instance, in a fully supervised manner, with clients training the target inference model for a predefined number of epochs, and after such a pre-training stage, the lower layers in the network might be used for mapping the input data points to the embedding space. The positive side of the former approach is that the encoder would not be tied to a particular classification and regression task, while the latter brings the benefit of the embeddings reflecting both the distribution over the target variables and the properties of the input data. Another benefit of using the target inference model is the fact, that there would be no need for training any additional model.

In \acronym, however, we also tested the appropriateness of an intermediate path, i.e. the \textit{supervised autoencoder} \cite{supervised_autoencoder}. This model architecture is an example of the multi-loss training paradigm in which multiple losses are optimized at the same time -- in our case, the classification or regression loss, and the reconstruction loss. The supervised autoencoder architecture consists therefore of \textit{i) }an encoder network, which takes the input data points and produces the embeddings, \textit{ii)} a classification or regression head, which takes the embeddings as produced by the encoder and uses them for predicting the target variables, and \textit{iii)} a decoder head, which takes the same embeddings and aims to reconstruct the input data points. The rationale behind this approach is that by introducing the reconstruction term, the embedding space is expected to better reflect the properties of the underlying features of the data points.

\section{Experimental Analysis}
\label{sec:results}

The goal of our experimental analysis is twofold. First, we want to confirm that \acronym client embeddings indeed reflect the actual data non-IIDness of federated datasets, and second, we want to assess the inference accuracy of clustered FL guided by \acronym.

\subsection{Experimental setup}

In our analysis, we focus on two types of non-IIDness that can be observed even in unlabeled data: the concept drift and the label skew described in the Preliminaries section. We investigate three datasets: \begin{enumerate*} [label=\itshape\alph*\upshape)] \item CIFAR10 \cite{cifar10}, for which we simulate the label skew by sampling data in a biased way (i.e. each client has a number of over- and other under-represented classes), and the concept drift by applying a different image augmentation pipeline on every client (e.g. on some clients the images are blurred, on others, the images are transformed to black and white, etc.); \item FEMNIST \cite{leaf_femnist}, which naturally contains the two considered non-IIDness types, and \item MNIST with pathological non-IIDness \cite{fl_seminal_paper}, which is obtained by partitioning the MNIST dataset \cite{deng2012mnist} in such a way that each of the artificially-generated clients contains images belonging to only two out of the ten available classes. \end{enumerate*}

Unless otherwise noted, the generated clients are randomly partitioned into two disjoint sets. The first set is the training clients $C_t$, which participate in the training procedure and therefore have both a training and validation set. The second set of clients $C_h$ is composed of the holdout clients, which only have a validation set. For the purpose of clustering, the training clients compute their client embeddings using their training dataset, while holdout clients compute it using their validation dataset.

Throughout this experimental section, we are going to be comparing the client embeddings as produced by the \acronym algorithm with the method proposed by \cite{clustering_hierarchical}, i.e. computing client embeddings as $e_i=\theta^T - \theta_i^{T+1}$, where $\theta_i^{T+1}$ is the model obtained after training the model $\theta^T$ on client $i$'s dataset $D_i$ for $f$ training epochs -- for conciseness, we refer to such an algorithm as WD (weight-difference). However, note that the WD method requires on-device training with a labeled dataset, thus, its use for computing client embeddings on holdout clients might not be possible in reality.

Regarding the model architecture, as a basis we use a simple encoder-decoder-classification head architecture with convolutional, ReLU, max pooling, and Softmax layers detailed in the Appendix. The encoder-classification head architecture is used for calculating WD's embeddings, while the encoder output is used for the \acronym-based clustering. In case of \acronym, the encoder is connected to either the classification head only (CLF), the decoder, thus forming an autoencoder (AE), or to both the classification head and the decoder, forming a supervised autoencoder (SAE). Unless stated otherwise, we use SAE.


\subsection{Correlation between Client Embeddings and Data-Generation Processes}

To estimate whether the embeddings reflect the actual non-IIDness, we first need to measure the similarity between two clients, i.e. the similarity of their underlying DGPs. Since we do not observe the DGPs directly we define the similarity between clients $i, j \in C $ as:
\begin{align}
    \mathcal{S}(i, j) \defeq \mathcal{S}(\gamma_i, \gamma_j) \approx \mathcal{S}(D_i, D_j).
\end{align}

How exactly is $\mathcal{S}(D_i, D_j)$ calculated depends on the nature of the non-IIDness. For the proposed CIFAR10 setting, the similarity may be computed as follows:
\begin{itemize}
    \item \textit{Label skew}: We first define the distribution vector $d_{i}$ as a vector in which every dimension $d_{i,k}\in \mathbb{N}_0$ states the number of data points in $D_i$ that belong to class $k$; then, the similarity between two datasets $D_i$ and $D_j$ can be estimated as the cosine similarity of the corresponding distribution vectors $d_i$ and $d_j$. That is, $\mathcal{S}(D_i, D_j)= \frac{d_i^T d_j}{\|d_i\|_2 \cdot\|d_j\|_2}$;
    \item \textit{Concept drift}: in our setting, the similarity of the two datasets $D_i$ and $ D_j$ is equal to the similarity of their clients' image augmentation pipelines. Such similarity can be estimated by applying the $i$'s and $j$'s augmentation pipelines to the same set of images, mapping the resulting images to an embedding space with a pre-trained network, and applying the softmax function to the so-obtained image embeddings in order to obtain valid probability distributions. As a final step, we compute the Jensen-Shannon divergence between every pair of probabilities that correspond to the same image, average them out, and then we convert the obtained JS divergence to a similarity with the formula $similarity=1-divergence$.
\end{itemize}

\begin{figure}[t]
\includegraphics[width=\columnwidth]{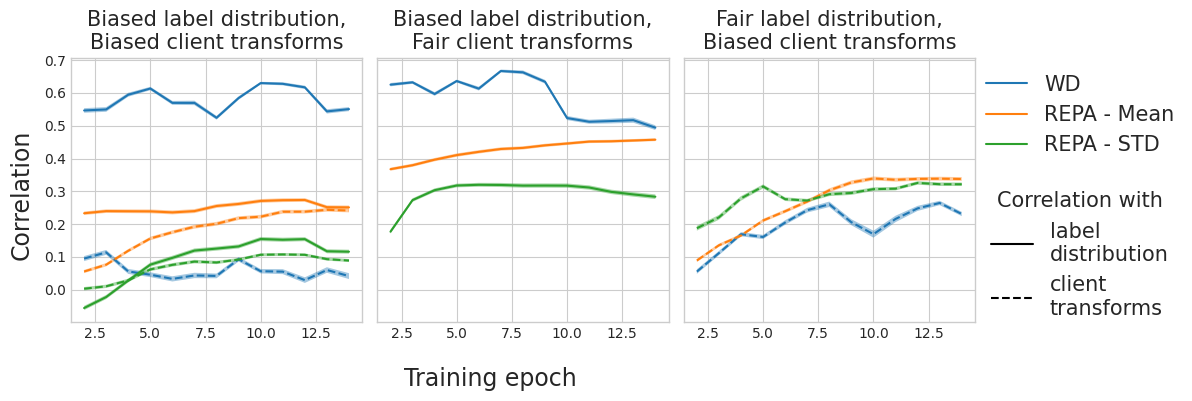}
\caption{Correlation of dataset similarities and the corresponding client embeddings obtained with WD and \acronym with client datasets sampled from CIFAR10.}
\label{fig:corr_wd}
\end{figure}

We set the number of training clients to $80$ and observe how the correlation between the similarity of clients $\mathcal{S}(i, j)$ and the cosine similarity of the client embeddings produced by the two algorithms under observation -- \acronym and WD -- evolve through the training on the CIFAR10 dataset. The results in \Cref{fig:corr_wd} show that WD is particularly successful at capturing the label skew type of non-IIDness as the correlation is, in some cases, even higher than $0.6$. This is unsurprising, as the algorithm relies on information stemming from local training, i.e. with local label (distributions) known. For \acronym we report the correlation of $\mathcal{S}$ with selected client embeddings statistics -- the mean and the quantiles (we used $0.25$, $0.5$, and $0.75$) -- that yield the best overall correlation. The figure demonstrates that, compared to WD, \acronym achieves a lower, but nevertheless, non-negligible correlation with the client similarity.


\subsection{Cluster Uniformity}

The goal of clustered FL is to bundle clients into groups exhibiting relative uniformity of the within-cluster clients' datasets. Thus, we now evaluate the ability of the \acronym clustering algorithm to generate uniform clusters. We focus on the MNIST dataset with pathological non-IIDness and define the uniformity of clusters $\mathbb{C}=\{c_k\}_{k=1}^{|\mathbb{C}|}$, where $c_k\subseteq C$ represents a set of clients in cluster $k$, as the average cosine similarity of the clustered clients' data distributions $d_i$s, i.e.:
\begin{align}
    U(C) &=\frac{1}{|\mathbb{C}|} \sum_{c\in \mathbb{C}} \frac{1}{{|c| \choose 2}} \sum_{i, j \in c, i\neq j}\frac{d_i^T d_j}{\|d_i\|_2\|d_j\|_2}.
\end{align}

\begin{figure}[t]
\includegraphics[width=\columnwidth]{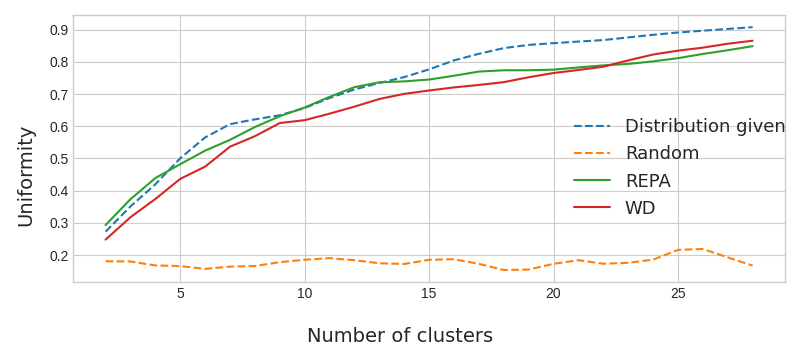}
\caption{Cluster uniformity when partitioning pathologically non-IID MNIST data.}
\label{fig:mnist_uniformity}
\end{figure}

In \Cref{fig:mnist_uniformity} we report the evolution of uniformity with respect to the number of clusters. We compare the values with the ones we get by randomly assigning the clients to one of the available clusters (Random) and the ones we get by clustering the distribution vectors directly with the K-Means clustering (Distribution given). We observe that \acronym clusters remain highly uniform, occasionally reaching the optimal uniformity. We also plot the performance of WD and conclude that training data-based clustering does not bring additional improvements.

\subsection{Cluster Robustness}

We argue that cluster robustness as defined in \Cref{def:cluster_robustness} is a key metric of practical importance in clustered FL:

\begin{definition}[Cluster robustness]
\label{def:cluster_robustness}
Cluster robustness is the probability that two clients with similar dataset properties belong to the same cluster.
\end{definition}

If we assume, that the cluster model $\theta_{\alpha(i)}$ is, among all the trained cluster models, the model that yields the best performance on dataset $D_i$, and that the performance of the cluster models $\theta_k$ do not deviate significantly over similar datasets, then cluster robustness states the probability that a previously unseen client is served the cluster model best suited for its local dataset. This probability cannot be measured directly, thus we estimate the robustness of a single cluster $r(c)$ empirically with \Cref{alg:robustness}. The robustness of the whole clustering structure $\mathbb{C}$ is then defined as $r(\mathbb{C})=\frac{1}{|\mathbb{C}|}\sum_{c\in \mathbb{C}} r(c)$. Note that in line 4 of the algorithm, we generate a client with a DGP similar to the DGP of an existing client whose data distribution is described with vector $d_i$. This is done by randomly sampling images so that the new client's data is distributed according to $(d_i+n)\cdot s$, where $n$ is a noise vector and $s$ is a scaling factor allowing the similar client to have a dataset of a different size. After sampling the images that compose the generated client's dataset, the reference client $r$'s image augmentation pipeline is applied to the generated dataset.

\begin{algorithm}[h!]
\caption{Algorithm for measuring cluster $c$'s robustness}
\label{alg:robustness}
\textbf{Input}: list of clients $L$ belonging to cluster $c$, $f_{emb}$ algorithm for computing client embeddings, clustering model $f_{clust}$\\
\textbf{Parameter}: number of iterations $n$\\
\textbf{Output}: Robustness of the cluster $c$

\begin{algorithmic}[1]
\STATE $n_s \gets 0$
\FOR{$i \gets 0$ to $n$}
    \STATE $r \gets$ sample client from $L$ uniformly at random
    \STATE $p \gets$ generate client with DGP similar as $r$ \label{line:similar_client}
    \STATE $e_p \gets f_{emb}(p)$ \COMMENT{compute client embedding}
    \STATE $c_p \gets f_{clust}(e_p)$ \COMMENT{determine cluster}
    \IF{$c_p = c$}
        \STATE $n_s \gets n_s+1$
    \ENDIF
\ENDFOR
\STATE robustness $\gets \frac{n_s}{n}$
\end{algorithmic}
\end{algorithm}

The robustness of the clusters obtained by the \acronym and WD is reported in \Cref{tab:robustness_by_noniidness}. \acronym manages to achieve overall robustness above 90\%, irrespective of the non-IIDness type. WD, on the other hand, struggles with the concept drift type of non-IIDness, where its robustness remains barely above the robustness that would have been achieved by randomly assigning clients to one of the $10$ available clusters. The table also shows that the supervised autoencoder (SAE) we embrace for \acronym outperforms alternative architectures.

\begin{table}[h]
\centering
\centerline{
\begin{tabular}{ll|rr|rr|rr}
\toprule
&  & \multicolumn{2}{c}{\thead{Concept drift,\\label skew}} & \multicolumn{2}{c}{\thead{Label skew}} & \multicolumn{2}{c}{\thead{Concept drift}} \\
&  & Mean & SE & Mean & SE & Mean & SE \\
\midrule
\multirow[c]{3}{*}{\acronym} & SAE & \textbf{0.91} & 0.02 & \textbf{0.90} & 0.02 & \textbf{0.92} & 0.02 \\
& CLF & 0.87 & 0.03 & \textbf{0.90} & 0.02 & 0.89 & 0.03 \\
& AE & 0.89 & 0.02 & 0.79 & 0.03 & 0.89 & 0.03 \\ \cline{2-8}
\multirow[c]{2}{*}{WD} & $f$=1 & 0.75 & 0.08 & 0.68 & 0.05 & 0.25 & 0.10 \\
& $f$=2 & 0.84 & 0.03 & 0.77 & 0.03 & 0.15 & 0.05 \\
\bottomrule
\end{tabular}
}
\caption{Robustness of clusters generated by \acronym and WD under various non-IIDness types. SAE stands for supervised autoencoder, AE for standard autoencoder, and CLF for classifier-based embeddings in \acronym, and $f$ is the number of fine-tuning epochs performed on the clients in case of WD. The number of clusters is in all cases set to 10.}
\label{tab:robustness_by_noniidness}
\end{table}

\subsection{Classification Accuracy}

We now assess the classification accuracy of the clustered FL models constructed by our \acronym approach as well as the alternative WD algorithm. As the baseline, we present the inference accuracy results achieved by the standard FedAvg approach. Throughout this section, we indicate with \textit{VAL} the classification accuracy (CA) of the model on the validation set of the training clients and with \textit{HO} the CA of the model on the validation set of the holdout clients. We conduct the evaluation on FEMNIST and MNIST with pathological non-IIDness, as these datasets naturally contain data distribution heterogeneity.

\subsubsection{FEMNIST dataset:} 
\begin{figure}[t]
\includegraphics[width=\columnwidth]{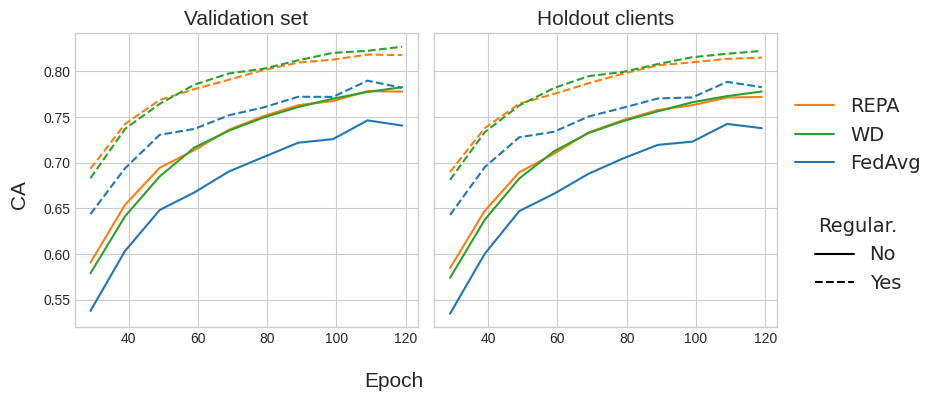}
\caption{Evolution of accuracy when training a model on the FEMNIST dataset with Federated Learning with respect to the algorithm used for computing the client embeddings. The number of clusters was set to $10$.}
\label{fig:acc_femnist}
\end{figure}

In \Cref{fig:acc_femnist} we see that \acronym and WD outperform the plain FedAvg approach on both VAL and HO, thus confirming the importance of client clustering when it comes to improving inference accuracy in FL over non-IID data. Note, however, that in reality the WD method cannot actually be used for computing the client embeddings of holdout clients, as it requires on-device training and a local labeled dataset, which is against our definition of the holdout client. We nevertheless report WD results here in order to assess a potential loss of inference accuracy in case a clustering approach such as \acronym is used on holdout clients that can actually perform the training. As evident from the figure, the loss is practically none.

As discussed in the related work, training loss regularization represents another means of improving FL performance in non-IID settings. Since the regularization remains orthogonal to the actual clustering scheme, we introduce the FedProx regularization term \cite{fedprox} during the individual client model training -- the loss $l(\theta_i)$ clients optimize over their datasets $D_i$ is updated to:
\begin{align}
    l_{reg}(\theta_i, \theta)=l(\theta_i) + \frac{\mu}{2}\|\theta_i - \theta\|_2^2,
\end{align}
where $\theta$ is the model sent by the server and $\mu$ the regularization strength. In \Cref{fig:acc_femnist} we see that such a regularization scheme significantly improves the inference accuracy, and hence in the remaining experiments we use FedProx regularization for all methods, unless stated otherwise.

\begin{figure}[t]
\includegraphics[width=\columnwidth]{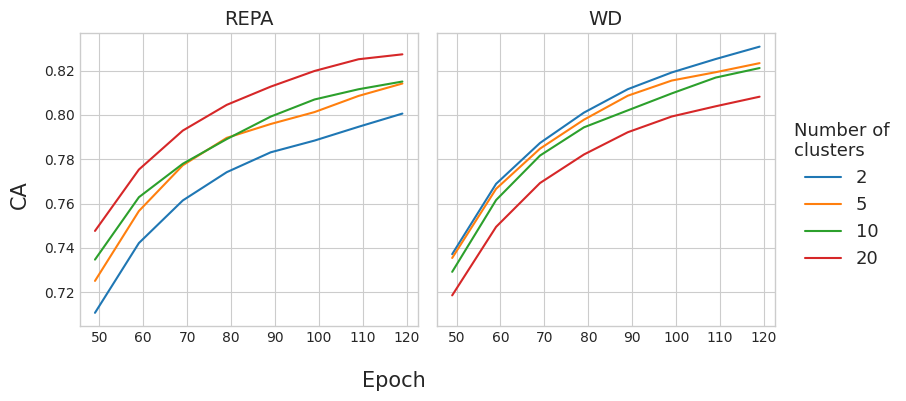}
\caption{Evolution of accuracy on holdout clients when training a model on the FEMNIST dataset with respect to the number of clusters.}
\label{fig:femnist_n_clusters}
\end{figure}

Finally, in \Cref{fig:femnist_n_clusters} we examine the importance of setting an appropriate number of clusters. In particular, we see that the WD algorithm achieves its highest accuracy when clustering the clients into $2$ clusters, while \acronym benefits from a higher number of clusters, such as $20$. This behavior can be in part explained by the results we reported for the robustness, as \acronym achieves significantly higher robustness than the WD algorithm. Note that, however, when clustering the clients there is a trade-off, as using a higher number of clusters results in increasingly uniform clusters, yet leads to fewer training clients within a cluster.

\subsubsection{MNIST with pathological non-IIDness:}

\begin{figure}[t]
\includegraphics[width=\columnwidth]{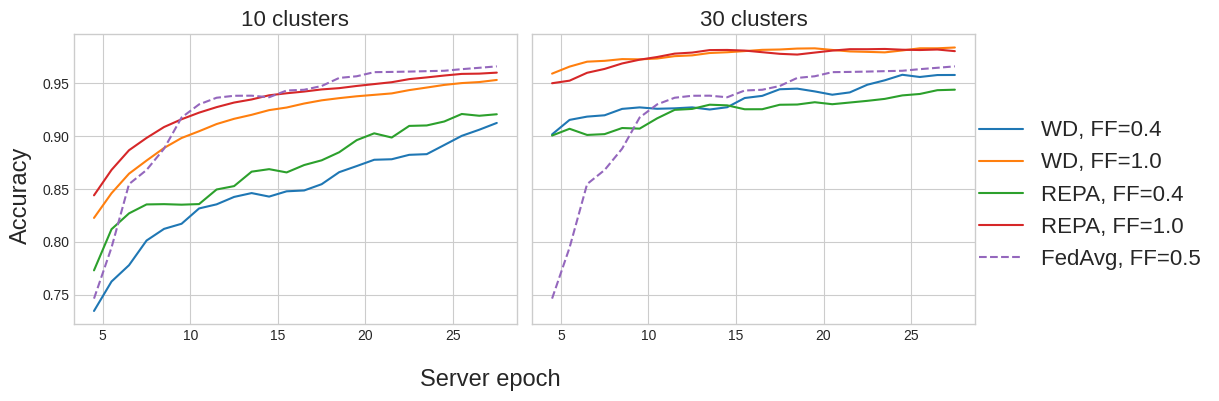}
\caption{Evolution of accuracy on the validation set of the training clients when training a model on the MNIST dataset with pathological non-IIDness. Models were trained with regularization. \textit{FF} stands for fraction fit, i.e. the percentage of clients that participate in every server training epoch. For \acronym, the model used to map the images to the embedding space is a classifier, and the mean and the standard deviation are computed over the embedding space.}
\label{fig:mnist_acc}
\end{figure}

Recall that in \Cref{fig:mnist_uniformity} we confirmed that both \acronym and WD manage to construct uniform clusters of clients. However, the increased uniformity alone does not guarantee higher inference accuracy, especially in highly non-IID settings. To demonstrate this, in \Cref{fig:mnist_acc} we plot the evolution of accuracy when $100$ clients  (we assume no holdout clients) are partitioned into $10$ and $30$ clusters using the two algorithms under observation. We see that, irrespective of the clustering algorithm used, the accuracy of clustered FL often does not improve over the accuracy of vanilla FL with FedAvg. This is due to two main reasons. First, because of the peculiarities of the pathological non-IID setting, the average accuracy suffers if not all clients participate in every server training epoch. If only a fraction of clients participate (e.g. fraction fit set to 0.4 in the graph), in clusters with more than two classes it is highly probable that in a given training epoch the cluster model does not get exposed to all the potential target labels. Simultaneously, as we measure the accuracy of the cluster model on all the clients in the cluster, the model is evaluated on clients containing data belonging to classes that might not have been present in the training set of the previous epoch. Therefore, in a certain sense, the model ``overfits'' the training dataset, and hence the quality of the predictions on the clients, which possess labels unseen in the previous training epoch, decreases. Second, compared to vanilla FL, in highly non-IID settings the negative effects of the decreased amount of data when building clustered models outweigh the benefits of increased homogeneity of clients used for model training, if the number of clusters is not sufficiently high (in this case 30).

\section{Discussion and Future Work}
\label{sec:discussion}

While the issue of poor model convergence and reduced inference accuracy due to non-IIDness has received substantial attention in the FL research community, the problem of assigning the most appropriate flavor of the joint model to clients without sufficiently large local datasets or to clients who merely want to use the join model has received, to the best of our knowledge, virtually no attention from the community. This is surprising, having in mind that in practice the majority of clients using FL-trained models do not participate in model training~\cite{bonawitz2019towards}. \acronym tackles this uncomfortable issue and includes a novel approach to client clustering that focuses on profiling the underlying data-generating processes. The existing clustering methods almost exclusively rely on clustering local models and their properties, which, we believe, is not sustainable in the long run, as it restricts model personalization to clients that participate in the training. As the popularity of FL grows, the number of clients who merely wish to use the models is likely to grow as well.


In this paper, we focused on devising and demonstrating \acronym. For practical applications, however, the clustering method would need to be optimized to the operating conditions. One of the main aspects of such optimization is the number of clusters in which the clients should be split. When determining this number we should aim to balance the homogeneity of individual clusters and the population of each cluster, as both the heterogeneity of the data within a cluster and the insufficient number of clients within a cluster may lead to poor model convergence. 
In future work, we plan on exploring various heuristics for determining the number of clusters within \acronym.

Finally, while alleviating the need for labeled data, \acronym still requires that FL clients collect a certain amount of unlabeled data for DGP profiling. In certain practical situations, even unlabeled data collection might not be possible, or the previously collected data might not accurately reflect the current DGP. This is especially true in mobile computing where a client might move to a different context (say, from a noisy outdoor area to a quiet indoor area), which might severely impact the underlying DGP. Nevertheless, contemporary mobile devices, such as smartphones, often host a range of embedded sensors that can be used to recognize context change. In forthcoming research, we intend to tap into this latent contextual data to infer information about the DGPs. This information, we believe, might enable client clustering even prior to data collection and, thus, further lower the entry bar for using personalized FL models.

\section{Conclusion}
\label{sec:conclusion}

In this paper we proposed \acronym, a client clustering scheme for federated learning that does not require labeled data, nor on-device training, yet enables clients to obtain model flavors tailored to the data distributions these clients observe. Extensive experimentation across different datasets demonstrates that \acronym brings inference accuracy comparable to that of the state-of-the-art FL model personalization efforts over a much wider range of clients, including those who did not participate in model training at all. Consequently, we believe that \acronym can facilitate a wider proliferation of federated learning, and can also serve as a basis for future efforts towards dynamic, context-aware FL adaptation.


\bibliography{aaai24}

\begin{thebibliography}{19}
\providecommand{\natexlab}[1]{#1}

\bibitem[{Bonawitz et~al.(2019)Bonawitz, Eichner, Grieskamp, Huba, Ingerman, Ivanov, Kiddon, Kone{\v{c}}n{\`y}, Mazzocchi, McMahan et~al.}]{bonawitz2019towards}
Bonawitz, K.; Eichner, H.; Grieskamp, W.; Huba, D.; Ingerman, A.; Ivanov, V.; Kiddon, C.; Kone{\v{c}}n{\`y}, J.; Mazzocchi, S.; McMahan, B.; et~al. 2019.
\newblock Towards federated learning at scale: System design.
\newblock In \emph{SysML (MLSys)}. Palo Alto, CA, USA.

\bibitem[{Briggs, Fan, and Andras(2020)}]{clustering_hierarchical}
Briggs, C.; Fan, Z.; and Andras, P. 2020.
\newblock Federated learning with hierarchical clustering of local updates to improve training on non-IID data.
\newblock In \emph{International Joint Conference on Neural Networks (IJCNN)}. Glasgow, UK.

\bibitem[{Caldas et~al.(2018)Caldas, Duddu, Wu, Li, Kone{\v{c}}n{\`y}, McMahan, Smith, and Talwalkar}]{leaf_femnist}
Caldas, S.; Duddu, S. M.~K.; Wu, P.; Li, T.; Kone{\v{c}}n{\`y}, J.; McMahan, H.~B.; Smith, V.; and Talwalkar, A. 2018.
\newblock Leaf: A benchmark for federated settings.
\newblock \emph{arXiv preprint arXiv:1812.01097}.

\bibitem[{Deng(2012)}]{deng2012mnist}
Deng, L. 2012.
\newblock The mnist database of handwritten digit images for machine learning research.
\newblock \emph{IEEE signal processing magazine}, 29(6): 141--142.

\bibitem[{Duan et~al.(2021)Duan, Liu, Ji, Liu, Liang, Chen, and Tan}]{duan2021fedgroup}
Duan, M.; Liu, D.; Ji, X.; Liu, R.; Liang, L.; Chen, X.; and Tan, Y. 2021.
\newblock Fedgroup: Efficient federated learning via decomposed similarity-based clustering.
\newblock In \emph{2021 IEEE Intl Conf on Parallel \& Distributed Processing with Applications, Big Data \& Cloud Computing, Sustainable Computing \& Communications, Social Computing \& Networking (ISPA/BDCloud/SocialCom/SustainCom)}. New York City, NY, USA.

\bibitem[{Ester et~al.(1996)Ester, Kriegel, Sander, Xu et~al.}]{dbscan}
Ester, M.; Kriegel, H.-P.; Sander, J.; Xu, X.; et~al. 1996.
\newblock A density-based algorithm for discovering clusters in large spatial databases with noise.
\newblock In \emph{KDD}. Portland, OR, USA.

\bibitem[{Ghosh et~al.(2020)Ghosh, Chung, Yin, and Ramchandran}]{clustering_send_all_2_all}
Ghosh, A.; Chung, J.; Yin, D.; and Ramchandran, K. 2020.
\newblock An efficient framework for clustered federated learning.
\newblock In \emph{NeurIPS}. Virtual.

\bibitem[{Huang et~al.(2019)Huang, Shea, Qian, Masurkar, Deng, and Liu}]{health_autoencoder}
Huang, L.; Shea, A.~L.; Qian, H.; Masurkar, A.; Deng, H.; and Liu, D. 2019.
\newblock Patient clustering improves efficiency of federated machine learning to predict mortality and hospital stay time using distributed electronic medical records.
\newblock \emph{Journal of biomedical informatics}, 99: 103291.

\bibitem[{Kairouz et~al.(2021)Kairouz, McMahan, Avent, Bellet, Bennis, Bhagoji, Bonawitz, Charles, Cormode, Cummings et~al.}]{loooong}
Kairouz, P.; McMahan, H.~B.; Avent, B.; Bellet, A.; Bennis, M.; Bhagoji, A.~N.; Bonawitz, K.; Charles, Z.; Cormode, G.; Cummings, R.; et~al. 2021.
\newblock Advances and open problems in federated learning.
\newblock \emph{Foundations and Trends{\textregistered} in Machine Learning}, 14(1--2): 1--210.

\bibitem[{Krizhevsky and Hinton(2009)}]{cifar10}
Krizhevsky, A.; and Hinton, G. 2009.
\newblock Learning multiple layers of features from tiny images.
\newblock Technical Report~0, University of Toronto.

\bibitem[{Le, Patterson, and White(2018)}]{supervised_autoencoder}
Le, L.; Patterson, A.; and White, M. 2018.
\newblock Supervised autoencoders: Improving generalization performance with unsupervised regularizers.
\newblock In \emph{NeurIPS}. Montreal, Canada.

\bibitem[{Li et~al.(2020)Li, Sahu, Zaheer, Sanjabi, Talwalkar, and Smith}]{fedprox}
Li, T.; Sahu, A.~K.; Zaheer, M.; Sanjabi, M.; Talwalkar, A.; and Smith, V. 2020.
\newblock Federated optimization in heterogeneous networks.
\newblock In \emph{MLSys}. Austin, TX, USA.

\bibitem[{Lloyd(1982)}]{kmeans}
Lloyd, S. 1982.
\newblock Least squares quantization in PCM.
\newblock \emph{IEEE transactions on information theory}, 28(2): 129--137.

\bibitem[{Long et~al.(2023)Long, Xie, Shen, Zhou, Wang, and Jiang}]{clustering_fesem}
Long, G.; Xie, M.; Shen, T.; Zhou, T.; Wang, X.; and Jiang, J. 2023.
\newblock Multi-center federated learning: clients clustering for better personalization.
\newblock \emph{World Wide Web}, 26(1): 481--500.

\bibitem[{McMahan et~al.(2017)McMahan, Moore, Ramage, Hampson, and y~Arcas}]{fl_seminal_paper}
McMahan, B.; Moore, E.; Ramage, D.; Hampson, S.; and y~Arcas, B.~A. 2017.
\newblock Communication-efficient learning of deep networks from decentralized data.
\newblock In \emph{International Conference on Artificial Intelligence and Statistics (AISTATS)}. Lauderdale, FL, USA.

\bibitem[{Mu et~al.(2023)Mu, Shen, Cheng, Geng, Fu, Zhang, and Zhang}]{fedproc}
Mu, X.; Shen, Y.; Cheng, K.; Geng, X.; Fu, J.; Zhang, T.; and Zhang, Z. 2023.
\newblock Fedproc: Prototypical contrastive federated learning on non-iid data.
\newblock \emph{Future Generation Computer Systems}, 143: 93--104.

\bibitem[{Ruan and Joe-Wong(2022)}]{ruan2022fedsoft}
Ruan, Y.; and Joe-Wong, C. 2022.
\newblock Fedsoft: Soft clustered federated learning with proximal local updating.
\newblock In \emph{AAAI}. Virtual.

\bibitem[{Sattler, M{\"u}ller, and Samek(2020)}]{clustered_fl_first}
Sattler, F.; M{\"u}ller, K.-R.; and Samek, W. 2020.
\newblock Clustered federated learning: Model-agnostic distributed multitask optimization under privacy constraints.
\newblock \emph{IEEE transactions on neural networks and learning systems}, 32(8): 3710--3722.

\bibitem[{Zhao et~al.(2018)Zhao, Li, Lai, Suda, Civin, and Chandra}]{zhao2018federatednoniid}
Zhao, Y.; Li, M.; Lai, L.; Suda, N.; Civin, D.; and Chandra, V. 2018.
\newblock Federated learning with non-iid data.
\newblock \emph{arXiv preprint arXiv:1806.00582}.

\end{thebibliography}

\end{document}